\pdfoutput=1

\documentclass[11pt]{article}

\usepackage[]{acl}

\usepackage{times}
\usepackage{latexsym}

\usepackage[T1]{fontenc}

\usepackage[utf8]{inputenc}

\usepackage{microtype}

\usepackage{inconsolata}
\usepackage{graphicx}
\usepackage{multirow}
\usepackage[table]{colortbl}
\usepackage{hhline}
\usepackage{amsmath}
\usepackage{tabularx}
\usepackage{xcolor}

\title{\textit{How You Prompt Matters!} Even Task-Oriented Constraints\\in Instructions Affect LLM-Generated Text Detection}

\author{Ryuto Koike$^1$ \quad Masahiro Kaneko$^{2,1}$ \quad Naoaki Okazaki$^1$ \\
  $^1$Tokyo Institute of Technology \quad $^2$MBZUAI\\
  \texttt{ryuto.koike@nlp.c.titech.ac.jp}\\\texttt{masahiro.kaneko@mbzuai.ac.ae} \quad \texttt{okazaki@c.titech.ac.jp}
  \\}

\begin{document}
\maketitle
\begin{abstract}
To combat the misuse of Large Language Models (LLMs), many recent studies have presented LLM-generated-text detectors with promising performance.
When users instruct LLMs to generate texts, the instruction can include different constraints depending on the user's need.
However, most recent studies do not cover such diverse instruction patterns when creating datasets for LLM detection.
In this paper, we reveal that even \textit{task-oriented} constraints --- constraints that would naturally be included in an instruction and are not related to detection-evasion --- cause existing powerful detectors to have a large variance in detection performance.
We focus on student essay writing as a realistic domain and manually create task-oriented constraints based on several factors for essay quality.
Our experiments show that the standard deviation (SD) of current detector performance on texts generated by an instruction with such a constraint is significantly larger (up to an SD of 14.4 F1-score) than that by generating texts multiple times or paraphrasing the instruction.
We also observe an overall trend where the constraints can make LLM detection more challenging than without them.
Finally, our analysis indicates that the high instruction-following ability of LLMs fosters the large impact of such constraints on detection performance.\footnote{Our detection datasets are available at \url{https://github.com/ryuryukke/HowYouPromptMatters}}
\end{abstract}

\begin{figure*}[t]
 \begin{center}
  \centering\includegraphics[width=0.96\textwidth]{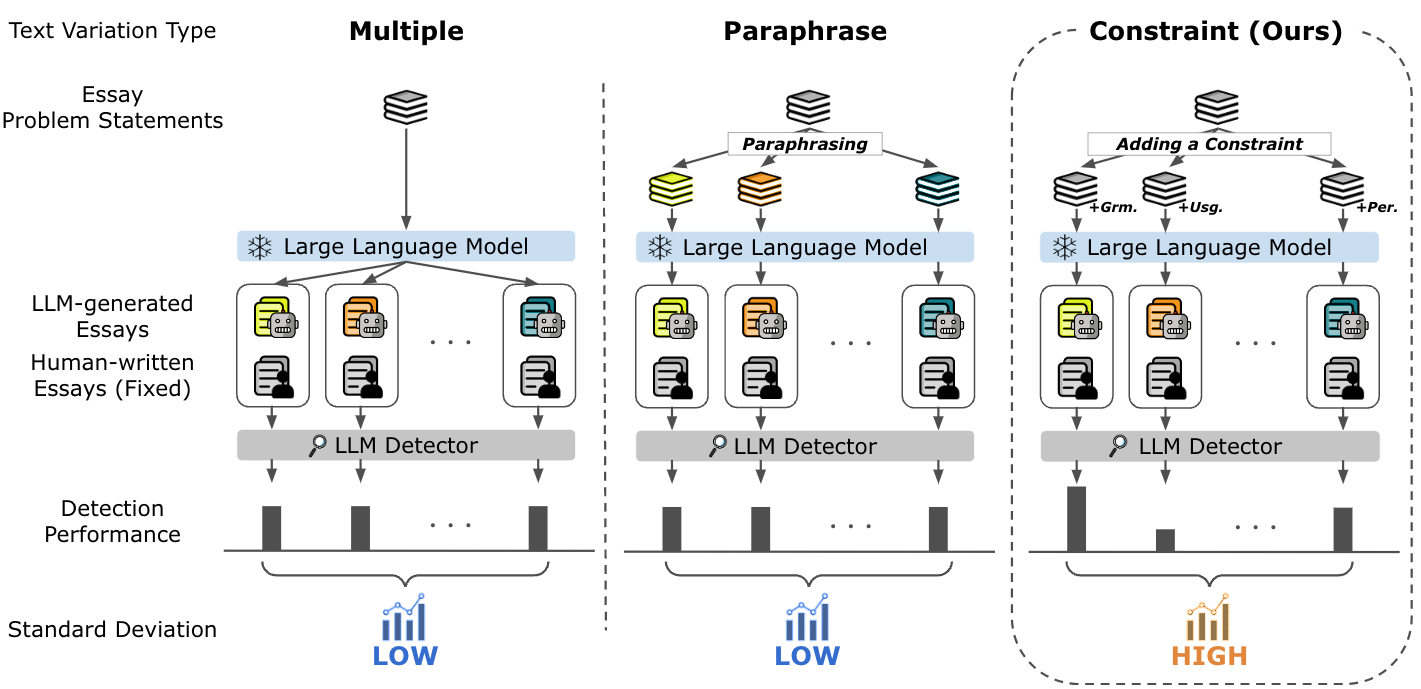}
  \caption{An overview of our \texttt{Constraint} setting and two baseline text variation types: \texttt{Multiple}, \texttt{Paraphrase}.
  To validate the impact of the constraint in instruction on LLM detection on the generated texts, we compare the SD of detection performance in our \texttt{Constraint} setting with that in two baseline settings: \texttt{Multiple}, \texttt{Paraphrase}.
  In the \texttt{Constraint} setting, \texttt{Grm.}, \texttt{Usg.}, and \texttt{Per.} are the abbreviation of factors for essay quality, listed in Table \ref{additional_constraints}.}
  \label{comparison}
 \end{center}
\end{figure*}

\begin{table*}[t]
    \centering
    \small
    \setlength{\tabcolsep}{8pt} 
    \renewcommand{\arraystretch}{1.2} 
    \begin{tabular}{lll}
    \hline
        \textbf{Factor} & & \textbf{Task-oriented constraint}\\\hline
        Grammatically & (Grm.) & Your essay must be free of grammatical errors.\\
        Usage & (Usg.) & Your essay must utilize a professional-level vocabulary.\\
        Mechanics & (Mec.) & Your essay must be free of spelling and capitalization errors.\\
        Style & (Sty.) & Your essay must include diverse word choices and sentence structures.\\
        Relevance & (Rel.) & Your essay must follow the prompt.\\
        Organization & (Org.) & Your essay must be logically organized.\\
        Development & (Dev.) & Your essay must include concrete evidence that supports your opinion.\\
        Cohesion & (Chs.) & Your essay must have a valid connection between paragraphs.\\
        Coherence & (Chr.) & Your essay must have an effective transition throughout all paragraphs.\\
        Thesis Clarity & (TC.) & Your essay must have a clear position through your essay.\\
        Persuasiveness & (Per.) & Your essay must be persuasive to readers.\\
    \hline
    \end{tabular}
    \caption{Task-oriented constraints for essay writing based on each factor of essay quality.}
    \label{additional_constraints}
\end{table*}

\section{Introduction}
LLMs have exhibited human-level generative capabilities in response to various textual instructions \cite{chatgpt,touvron2023llama}.
With such remarkable generative ability, malicious users might exploit LLMs for cheating on student homework or fabricating misinformation \cite{tang2023science,wu2023survey}.
To mitigate such potential misuse of LLMs, many recent works have presented LLM-generated-text detectors with highly promising detection performance \cite{kirchenbauer2023watermark,mitchell2023detectgpt,guo2023close,Koike:OUTFOX:2024,su2023detectllm}.

When users instruct LLMs to generate texts, the instruction potentially includes various constraints (e.g., output format and style) \cite{prompt2023engineering}.
Here, we call such constraints that would naturally be included in instruction and are not related to detection-evasion as \textit{task-oriented} constraints.
Despite being very natural, such differences in the instruction can have a large impact on the quality of the generated texts or on the downstream performance of various NLP tasks \citep{jiang2020know,zhang2023prompting,feng2023sentence}.
Most studies in LLM detection focus on the target LLM-generated text itself, analyzing its linguistic features \citep{mitrović2023chatgpt,li2023deepfake,guo2023close,liu2023argugpt} and not how the target texts are generated.
Moreover, most previous works do not include such a variety of instructions to create their benchmarking datasets for LLM detection \citep{li2023deepfake,guo2023close,liu2023argugpt}. 
This paper sheds light on the following question: \textit{\textbf{Do the task-oriented constraints in generation instruction affect the LLM detection?}}

Motivated by this question, this paper first demonstrates that even task-oriented constraints in instruction can lead to inconsistent detection performance of current significant detectors.
In particular, as depicted on the right in Figure \ref{comparison}, we explore the standard deviation (SD) of detection performance on various datasets generated via instructions with each different constraint.
We focus on student essay writing as one of the generation tasks to consider the constraints, and there is a recognized demand for its detection \cite{educatorconsiderations4chatgpt}.
To generate essays via LLMs, we utilize essay questions created by \citet{Koike:OUTFOX:2024}.
Then, as listed in Table \ref{additional_constraints}, we manually create the task-oriented constraints based on each factor of essay quality, defined by \citet{KeN19}.
To verify the impact of the constraint, we compare the SD with that of two baseline text variation types: generating texts multiple times (via sampling) and paraphrasing the instruction, denoted as \texttt{Multiple} and \texttt{Paraphrase} in Figure \ref{comparison}.

Indeed, our experiments show that a task-oriented constraint in the instruction has a more significant effect on the detection performance than the randomness caused by sampling texts or paraphrasing the instruction.
Specifically, the SD of current detector performance on texts generated in our \texttt{Constraint} setting is substantially higher (up to an SD of 14.4 F1-score) than that in the \texttt{Multiple} and \texttt{Paraphrase} settings.
We also observe an overall trend where the constraints can make LLM detection more challenging than without them.
Finally, our analysis suggests that the high instruction-following ability of LLMs causes the large impact of such constraints on detection performance.

\begin{figure*}[t]
 \begin{center}
  \centering\includegraphics[width=0.94\textwidth]{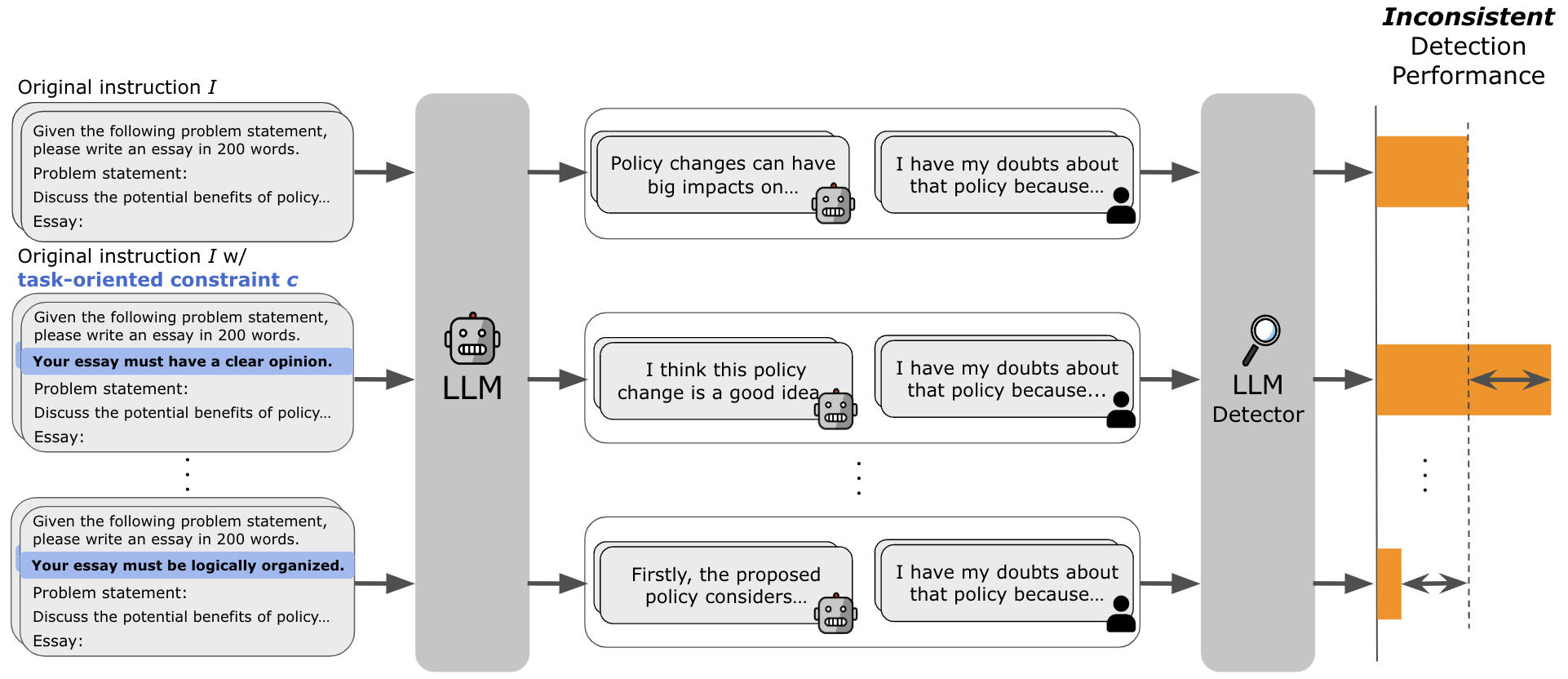}
  \caption{Even a task-oriented constraint in the instruction can cause inconsistent detection performance on the LLM-generated texts.}
  \label{fig:overall}
 \end{center}
\end{figure*}

\section{Methodology}
\label{method}
This section describes our strategy for identifying and evaluating the impact of task-oriented constraints in the instruction on the performance of LLM-generated text detection.

\subsection{Task Formulation}
\label{task_formulation}
Our main task is LLM-generated text detection, specifically discerning LLM-generated essays from human-written essays.
To evaluate the performance of current detectors, we utilize a mixture of human-written and LLM-generated essays as our test set.
We employ pairs of essay problem statements $s_j$ and human-written essays $h_j$ that are part of the essay dataset created by \citet{Koike:OUTFOX:2024}.
Then, we instruct LLMs to generate essays based on the problem statements $s_j$.
We elaborate on the details of our test set in \S\ref{setup}.

\subsection{Investigating the Impact of Task-Oriented Constraints on LLM Detection}
\label{all_setting}
In our work, we investigate the variation in the detection performance of texts generated by instruction with different constraints.
\paragraph{Constrained Generation}
We instruct an LLM to generate an essay and obtain a set of essays for each different constraint included in the instruction.
Let $I_{i}$ an instruction including the task-oriented constraint $c_{i}$, we instruct an LLM to generate an essay $e_{ij}$ based on an essay problem statement $s_{j}$,
\begin{equation}
e_{ij} = LLM\left(I_{i}, s_{j}\right) .
\end{equation}

\noindent
To facilitate our study, we manually\footnote{We create our task-oriented constraints as simply as possible to explain the factor, following the "Start Simple" philosophy in prompt engineering \cite{prompt2024engineering}.} create the constraints $c_{i}$ on essay writing based on each factor of essay quality, defined by \citet{KeN19}.
Table \ref{additional_constraints} lists the factors and our constraints.
$I_i \left(i = 1\right)$ is the original instruction, not including any constraints:

\noindent
\begin{verbatim}
Given the following problem statement,
please write an essay in {n} words.
Problem statement:
{problem_statement}
Essay: 
\end{verbatim}

\noindent
where \texttt{\{n\}} is the number of words in a human-written essay $h_j$ paired with an essay problem statement $s_j$ and \texttt{\{problem\_statement\}} denotes the essay problem statement $s_j$.

$I_i \left(2 \leq i \leq 12\right)$ is the original instruction with an added constraint from Table \ref{additional_constraints}:

\noindent
\begin{verbatim}
Given the following problem statement,
please write an essay in {n} words.
{constraint_i}
Problem statement:
{problem_statement}
Essay: 
\end{verbatim}

\noindent
where \texttt{\{n\}} and \texttt{\{problem\_statement\}} are the same as the above and \texttt{\{constraint\_i\}} denotes the constraint $c_i$ in Table \ref{additional_constraints}. For instance, ``Your essay must be logically organized.''

\paragraph{Impact Evaluation}
As depicted in Figure \ref{fig:overall}, we leverage LLMs to generate essays via instructions without and with each different constraint $c_i$ and thus obtain the original instruction dataset and multiple constraint-based datasets $DS_i = \{\left(h_j, e_{ij}\right)\}_{j=1}^{N}$.
To quantify the impact of the constraint on LLM detection, let $p_i$ be the F1-score detection performance on $DS_i$, we calculate the SD of the detection performance on the multiple datasets,
\begin{equation}
\sigma = \sqrt{\frac{\sum_{i=1}^{12}\left(p_{i}-\mu\right)^2}{12}} .
\end{equation}

\noindent
Here, $\mu$ is the average of the detection performances $\{p_i\}_{i=1}^{12}$.
To validate the impact of such constraints on LLM detection, we use two randomnesses as baseline text variation types: \texttt{Multiple} and \texttt{Paraphrase}.
Finally, we compare the SD of the detection performance in our \texttt{Constraint} setting of adding a constraint to instruction and \texttt{Multiple} and \texttt{Paraphrase}.
We delve into the \texttt{Multiple} and \texttt{Paraphrase} settings in \S\ref{baaseline_variation_type}.

\section{Experiments and Results}
\label{experiment}
Our experiment investigates the answer to the following question: \textit{Can current detectors consistently capture LLM-generated text variations caused by even task-oriented constraints in the instruction?}

\subsection{Experimental Setup}
\label{setup}
\paragraph{Essay Generation Models}
We employ ChatGPT (gpt-3.5-turbo-0613) and GPT-4 (gpt-4-0613), which are commonly used LLMs, as our essay generation model.
Additional configuration details of the essay generator models are in Appendix \ref{config_details}.

\paragraph{Evaluation Metric and Dataset}
In our experiment, as described below, all LLM-generated-text detectors output a binary label for an input text. Thus, our evaluation metric for LLM detectors is the F1-score on LLM-generated texts, which is a common evaluation metric in binary classification tasks.
As our evaluation dataset, we employ pairs of essay problem statements and human-written essays from the essay dataset created by \citet{Koike:OUTFOX:2024}.
We also prepare a set of LLM-generated essays based on the same essay problem statements.
Finally, we evaluate LLM detectors on a mixture of 500 human-written and 500 LLM-generated essays from the problem statements by each LLM.

\paragraph{LLM-generated Text Detectors}
To verify the stability of existing representative detectors\footnote{The logit information of our essay generators (ChatGPT and GPT-4) is not publicly available, thus our study does not cover statistical outlier detectors.}, we target the HC3 ChatGPT detector\footnote{\url{https://huggingface.co/Hello-SimpleAI/chatgpt-detector-roberta}} and the ArguGPT\footnote{\url{https://huggingface.co/SJTU-CL/RoBERTa-large-ArguGPT}} as supervised classifiers, and the in-context learning (ICL) approach\footnote{\url{https://github.com/ryuryukke/OUTFOX}} of \citet{Koike:OUTFOX:2024}.
  The HC3 detector is a RoBERTa-base detector fine-tuned with the Human ChatGPT Comparison Corpus (HC3) dataset for detecting ChatGPT-generated texts covering diverse domains \citep{guo2023close}.
  The ArguGPT is a RoBERTa-large detector fine-tuned for catching LLM-generated argumentative essays, including various domains such as homework exercises, TOEFL, and GRE writing tasks \citep{liu2023argugpt}.
  Following the setting of \citet{Koike:OUTFOX:2024}, in the ICL approach, we leverage ChatGPT (gpt-3.5-turbo-0613) with 5 ChatGPT-generated and 5 human-written essays as in-context examples from their training set for each essay to be detected. 
  Further configuration details of the in-context learning approach are in Appendix \ref{config_details}.

\paragraph{Text Variation Type}
\label{baaseline_variation_type}
In our experiment, to verify the consequent impact of a task-oriented constraint on LLM detection, we have two text variation types as the baseline: \texttt{Multiple} and \texttt{Paraphrase}, which can influence the stability of detection performance.
\begin{itemize}
  \item \textbf{\texttt{Multiple}:} 
  As depicted in Figure \ref{additional_constraints}, 
  we instruct LLMs to generate 12 texts from each original instruction to form 12 datasets.
  To generate multiple texts from an instruction, we utilize an argument to control the number of outputs in the OpenAI Chat Completion API\footnote{\url{https://platform.openai.com/docs/api-reference/chat/create##chat-create-n}}.
  The SD of detection performance is calculated on the 12 datasets. 
  \item \textbf{\texttt{Paraphrase}:} 
  As shown in Figure \ref{additional_constraints}, we obtain 12 datasets by generating a text from each of 12 different paraphrases of the original instruction. The SD of detection performance is computed on the 12 datasets. 
  To paraphrase the original instruction, we employ ChatGPT (gpt-3.5-0613).
  Specifically, we paraphrase the beginning of the original instruction thus, the instruction in this setting is as follows,
\begin{verbatim}
{paraphrased_instruction}
Problem statement:
{problem_statement}
Essay: 
\end{verbatim}

  \noindent
  where \texttt{\{paraphrased\_instruction\}} is a paraphrase of the beginning of the original instruction, which is ``Given the following problem statement, please write an essay in \texttt{\{n\}} words.''.
  For instance, \texttt{\{paraphrased\_instruction\}} can be ``I kindly request you to compose an essay that adheres to the given problem statement, ensuring that it contains \texttt{\{n\}} words.''.
  The examples of the paraphrases are in Appendix \ref{config_details}.
\end{itemize}

\begingroup
\begin{table}[t]
\centering
\small
\setlength{\tabcolsep}{4pt} 
\renewcommand{\arraystretch}{1.25} 
\begin{tabular}{cccccc}\hline
\multirow{2}{*}{\textbf{\shortstack{Essay\\Generator}}} & \multirow{2}{*}{\textbf{\shortstack{Variation Type}}} & \multicolumn{3}{c}{\textbf{LLM Detector}}\\\cline{3-5}
 & & \textbf{HC3} & \textbf{ArguGPT} & \textbf{ICL} \\\hline
\multirow{14}{*}{ChatGPT}& Multiple & 1.02 & 0.30 & 0.48 \\
& Paraphrase & 4.07 & 0.84 & 0.58 \\
& Constraint (Ours) & \textbf{12.76} & \textbf{6.69} & \textbf{1.15} \\
\hhline{~====}
& \multicolumn{4}{c}{\textbf{The \textit{deviation} for each factor}}\\
& \multicolumn{1}{l}{Grammar} & \multicolumn{1}{c}{5.43} & 2.43 &  0.13 \\
& \multicolumn{1}{l}{Usage} & \multicolumn{1}{c}{34.78} & 19.77 &  1.73 \\
& \multicolumn{1}{l}{Mechanics} & \multicolumn{1}{c}{2.23} &  1.33 &  0.98 \\
& \multicolumn{1}{l}{Style} & \multicolumn{1}{c}{15.88} &  5.67 &  0.83 \\
& \multicolumn{1}{l}{Relevance} & \multicolumn{1}{c}{5.13} &  2.53 &  0.23 \\
& \multicolumn{1}{l}{Organized} & \multicolumn{1}{c}{6.23} &  2.83 & 1.03  \\
& \multicolumn{1}{l}{Development} & \multicolumn{1}{c}{3.33} & 3.33  &  1.98 \\
& \multicolumn{1}{l}{Cohesion} & \multicolumn{1}{c}{11.23} &  3.43 &  0.63 \\
& \multicolumn{1}{l}{Coherence} & \multicolumn{1}{c}{7.03} & 2.83  &  0.03 \\
& \multicolumn{1}{l}{Thesis Clarity} & \multicolumn{1}{c}{4.03} &  2.43 &  2.08 \\
& \multicolumn{1}{l}{Persuasive} & \multicolumn{1}{c}{0.53} &  1.63 &  0.23 \\
\hline
\multirow{14}{*}{GPT-4}& Multiple & 1.09 & 1.14 & 0.68 \\
& Paraphrase & 3.42 & 2.43 & 0.69 \\
& Constraint (Ours) & \textbf{4.13} & \textbf{14.38} & \textbf{1.26}\\
\hhline{~====}
& \multicolumn{4}{c}{\textbf{The \textit{deviation} for each factor}}\\
& \multicolumn{1}{l}{Grammar} & 2.26  &  8.11 &  0.85 \\
& \multicolumn{1}{l}{Usage} & 8.34  &  34.39 &  1.75 \\
& \multicolumn{1}{l}{Mechanics} &  2.96 &  8.01 &  0.45 \\
& \multicolumn{1}{l}{Style} &  7.54 &  24.39 &  0.55 \\
& \multicolumn{1}{l}{Relevance} &  1.96 &  7.01 &  0.15 \\
& \multicolumn{1}{l}{Organized} &  4.26 &  7.01 &  0.05 \\
& \multicolumn{1}{l}{Development} &  0.44 &  9.61 &  0.35 \\
& \multicolumn{1}{l}{Cohesion} &  3.56 &  4.81 &  0.05 \\
& \multicolumn{1}{l}{Coherence} &  1.44 &  4.89 &  1.05 \\
& \multicolumn{1}{l}{Thesis Clarity} &  0.26 &  6.21 &  1.05 \\
& \multicolumn{1}{l}{Persuasive} &  0.74 &  4.21 &  3.25 \\
\hline
\end{tabular}
\caption{\label{table: main_result}
A comparison of the SD of detection performance on essays generated by ChatGPT and GPT-4 on three variation types: \texttt{Multiple}, \texttt{Paraphrase}, and \texttt{Constraint}.
It includes the \textit{deviation} of detection performance for each factor in our \texttt{Constraint} setting.
}
\end{table}
\endgroup

\subsection{Results}
\label{main_result}
Table \ref{table: main_result} presents the comparison of the SD of detection performance in the three text variation settings: \texttt{Multiple}, \texttt{Paraphrase}, and \texttt{Constraint}. 
In addition, it shows the deviation of detection performance for each factor in our \texttt{Constraint} setting.
The LLM detectors include the HC3 detector, ArguGPT, and the ICL approach.
Throughout all configurations of the generator and the detector, the SD on texts via instruction with the constraints is significantly larger than the two baseline variation types, reaching up to an SD of 14.4 F1-score for ArguGPT.
These results imply that even a task-oriented constraint in instruction has a more significant effect on the detection performance of current detectors than the effect of generating texts multiple times and paraphrasing the instruction.
We also observe an overall trend where the constraints make the detection more challenging: there is a decrease in detection performance in most constraints with up to a 40.3 drop in F1-score. We provide the detection performance itself in Appendix \ref{appendix_b}.

Especially in the HC3 detector and ArguGPT, we can observe that the SD of detection performance is relatively large. This may be partially because the two detectors are trained with benchmarking datasets created without considering a variety of instructions and are prone to the difference of constraint in instruction.
On the other hand, in the ICL approach, the effect of the constraint in instruction is relatively small. We could assume that the ICL approach might inherently consider a wide variety of expressions as in-context examples for detection, alleviating the effect of the constraint.

Looking into the impact of constraint for each factor, throughout all settings of the generator and the detector, the deviation of detection performance is notably larger in the factors of ``Usage'' and ``Style''.
As shown in Table \ref{additional_constraints}, since both constraints on the two factors explicitly instruct to change the lexical distribution of the output text, this result aligns with our expectation.
In our pilot study, we calculate the average of distinct-n $\left(=1, 2, 3\right)$\footnote{The distinct-n is a metric for expression diversity in multiple texts, calculating the ratio of unique n-grams in the total word count. We apply the distinct-n to the setting of two texts to measure the difference in expression between them.} \cite{zhao-etal-2017-learning,li-etal-2016-diversity} between two output texts from instructions without and with constraints for each factor through our test set. As a result, the top two factors with distinct-n values are ``Usage'' and ``Style''.
This implies that the constraints of ``Usage'' and ``Style'' in instruction may cause relatively large differences in the expression of the output texts, leading to such a large impact on the detection performance.


\begingroup
\begin{table*}[t]
\centering
\small
\setlength{\tabcolsep}{6pt} 
\renewcommand{\arraystretch}{1.3} 
\begin{tabular}{ccccccc}\hline
\multirow{2}{*}{\textbf{Essay Generator}} & \multirow{2}{*}{\textbf{Text Variation Type}} & \multicolumn{3}{c}{\textbf{LLM Detector}} & 
 \multirow{2}{*}{\textbf{\shortstack{Instruction-Following\\Score (\%)}}} \\\cline{3-5}
 & & \textbf{HC3} & \textbf{ArguGPT} & \textbf{ICL} \\\hline
\multirow{3}{*}{ChatGPT}& Multiple & 1.02 & 0.30 & 0.48 & \multirow{6}{*}{\textbf{87.1}} \\
& Paraphrase & 4.07 & 0.84 & 0.58 & \\
& Constraint & \textbf{12.76} & \textbf{6.69} & \textbf{1.15} & \\
\cline{1-5}
\multirow{3}{*}{GPT-4}& Multiple & 1.09 & 1.14 & 0.68 & \\
& Paraphrase & 3.42 & 2.43 & 0.69 & \\
& Constraint & \textbf{4.13} & \textbf{14.38} & \textbf{1.26} & \\
\hline
\multirow{3}{*}{Davinci-002}& Multiple & 1.07 & 0.15 & 0.78 & \multirow{3}{*}{49.3} \\
& Paraphrase & \textbf{4.14} & \textbf{0.51} & \textbf{1.51} &  \\
& Constraint & 1.44 & 0.32 & 1.17 & \\
\hline
\end{tabular}
\caption{\label{davinci_result}
A comparison of the SD of detection performance on essays generated by ChatGPT, GPT-4, and Davinci-002 on three variation types: \texttt{Multiple}, \texttt{Paraphrase}, and \texttt{Constraint} (as described in \S\ref{setup}). The instruction-following score is a ratio of texts that follow a constraint in the generated texts by instruction with a constraint. It shows the overall instruction-following score across all constraints.}
\end{table*}
\endgroup

\begin{table*}
    \centering
    \small
    \setlength{\tabcolsep}{5pt} 
    \renewcommand{\arraystretch}{1.5} 
    \begin{tabular}{ccccccccccccc}
    \hline
     \multirow{2}{*}{\textbf{Essay Generator}} & \multicolumn{11}{c}{\textbf{Factor}} & \\\cline{2-12}
     & \textbf{Chs.} & \textbf{Chr.} & \textbf{Org.} & \textbf{Rel.} & \textbf{Sty.} & \textbf{Usg.} & \textbf{Dev.} & \textbf{TC.} & \textbf{Per.} & \textbf{Grm.} & \textbf{Mec.} & \textbf{Overall}\\\hline
     ChatGPT/GPT-4 & 100 & 100 & 100 & 95.6 & 93.3 & 91.1 & 88.9 & 88.9 & 84.4 & 71.1 & 44.4 & 87.1 \\\hline
    \end{tabular}
    \caption{The ratio of essays that follow each constraint in a mixture of essays generated by ChatGPT and GPT-4 with the instruction including each constraint. The scores are sorted in descending order.}
    \label{instruction-following-for-each-factor}
\end{table*}

\section{High Instruction-Following Ability Leads to Inconsistent Detection}
\label{analysis}
Our experiments demonstrate that even task-oriented constraints in instruction induce notably inconsistent detection performance of the generated texts.
In this section, to further examine how the constraint causes such an effect, we verify our hypothesis: \textit{The high instruction-following ability of LLMs as a generator causes the large impact of constraints in the instruction on LLM detection}.


\subsection{Verification Setup}
\label{analysis_gpt4}
To verify our hypothesis, we investigate the relationship between the level of instruction-following ability of a generator and the extent of the impact of a constraint.
In particular, we compare the impact of a constraint when using generators with low and high instruction-following ability as we measure.

\paragraph{Evaluating the Instruction-Following Ability}
To evaluate the instruction-following ability of an LLM, we calculate the \textit{instruction-following score}: the ratio of texts that follow a constraint in generated texts by instruction with a constraint.
Here, we call the texts generated by instruction without and with constraint as \textit{plain} and \textit{constrained} texts each.
We prompt GPT-4\footnote{GPT-4 as an evaluator has been recently reported to exhibit promising alignment with humans in evaluation across various downstream tasks \cite{liu2023geval,chiang-lee-2023-large}.} to classify whether a constrained text follows the constraint or not, compared with a plain text,

\begin{verbatim}
Please classify whether the following
texts follow the constraint.
Constraint: {constraint}
Text: {plain_text}
Answer (just Yes or No): No
Text: {constrained_text}
Answer (just Yes or No):
\end{verbatim}

\noindent
where \texttt{\{constraint\}} is a task-oriented constraint we create, for instance including ``Your essay must be logically organized.'' and \texttt{\{plain\_text\}} and \texttt{\{constrained\_text\}} denotes a plain text and constrained text based on the same essay problem statement, respectively. 
To eliminate the randomness of the evaluation by GPT-4, we configure $\mathsf{temperature}$ and $\mathsf{top\_p}$ parameters to be 0.

For 11 task-oriented constraints each, we sample 45 pairs of plain text and constrained text from our test set generated by an LLM. Finally, we compute the instruction-following score on 495 $\left(=45\times11\right)$ texts generated by an LLM.

\paragraph{Comparing the Impacts of a Constraint between LLMs}
Besides ChatGPT and GPT-4, we explore the effect of a constraint when using an LLM with relatively low instruction-following ability as a generator.
As such a generator model with low instruction-following ability, we select Davinci-002 (GPT-3), which is not explicitly optimized for instruction-following.\footnote{OpenAI declares that GPT base models, including Davinci-002, are not optimized for instruction-following: \url{http://tinyurl.com/gpt-base-model}.}
The parameter configuration of Davinci-002 is in Appendix \ref{config_details}.


\begin{table*}[ht]
  \small
  \setlength{\tabcolsep}{4.5pt}
  \renewcommand{\arraystretch}{1.3}
  \centering
  \begin{tabularx}{\textwidth}{p{3.2cm}X}
    \hline
    \multicolumn{1}{l}{Essay Problem Statement} & \texttt{Should schools require a minimum grade requirement for students to participate in sports? Discuss the potential benefits and drawbacks of such a policy, taking into consideration the impact on students' academic performance, physical health, and personal interests. Use evidence and examples to support your argument.} \\
    \hline
    \multicolumn{1}{l}{Plain Essay} & \texttt{Education and sports are two integral parts of a student's life. However, the debatable question regarding the minimum grade for students to participate in sports has been a contentious topic of discussion. Supporters of the minimum grade requirement argue that it presents students with a target to work towards, thus promoting focus and dedication towards their academic performance. With this policy in place, students are motivated to perform well acadically because participation in sports depends on it. This, in turn, helps them develop better time management and {\color{red}{prioritization}} skills. Moreover, this prerequisite can drive students to have greater self-discipline, which} \\\hline
  \multicolumn{1}{l}{Constrained Essay} & \texttt{Educational institutions universally grapple with the {\color{red}{conundrum}} of whether to impose a minimum grade requirement for students wishing to partake in sports. While the underlying purpose of such a provision is to ensure academic excellence, it may inadvertently suppress the sporting interests of students, thus bringing their physical health into question. The most salient benefit of imposing a minimum grade requirement is undoubtedly the enhancement of academic performance. By establishing a tangible prerequisite for sports participation, it {\color{red}{incentivizes}} students to strive for scholastic excellence. Additionally, this measure can circumvent the {\color{red}{pervasiveness}} of {\color{red}{overemphasis}} on athletics sometimes observed in schools, inadvertently}\\
    \hline
  \end{tabularx}
  \caption{Examples of essays generated by GPT-4 via instruction without and with a constraint on ``Usage''. Both essays are cut off at the first 100 words. {\color{red}{Red}} in the table are low-frequency words based on the Wikipedia dump.}
  \label{tab:constrained_generation}
\end{table*}

\subsection{Results}
\paragraph{Overall}
Table \ref{davinci_result} provides a comparison of the SD of detection performance on essays generated by ChatGPT, GPT-4, and Davinci-002, in the three variation settings: \texttt{Multiple}, \texttt{Paraphrase}, and \texttt{Constraint}.
It also includes the instruction-following score of Davinci-002 essays and a mixture of essays generated by ChatGPT and GPT-4, each across all constraints\footnote{We group LLMs based on their instruction-following ability and treated ChatGPT and GPT-4 as a relatively high-performing group compared with Davinci-002.}.

We can observe that the effect of a constraint in the case of ChatGPT and GPT-4 is significantly large, but the effect in the case of Davinci-002 is quite small. Furthermore, the instruction-following score in a group of ChatGPT and GPT-4 across all factors is 87.1\%, which is notably larger than 49.3\% in Davinci-002.\footnote{To confirm the validity of the evaluation by GPT-4, we use Amazon Mechanical Turk to crowdsource the human agreement rate with the evaluation. We get an 87\% human agreement rate, ensuring the validity of the evaluation to some degree. The details of this validation are in Appendix \ref{human_agreement_details}.}
These results imply that the high instruction-following ability of LLMs reinforces the effect of a constraint on LLM detection, supporting our hypothesis. We provide a discussion on the change in detection difficulty caused by the constraints, showing the detection performance itself in Appendix \ref{appendix_b}.

\paragraph{Details}
Table \ref{instruction-following-for-each-factor} shows the instruction-following score for each constraint in a group of ChatGPT and GPT-4. We sort the scores in descending order.
Here, the top three constraints with the deviation in detection performance, averaged between the detectors and essay generators, are ``Usage'', ``Style'', and ``Cohesion'', while the bottom three constraints are ``Persuasive'', ``Mechanics'', and ``Thesis Clarity''.
All top three constraints obtain over 90\% of the instruction-following score and are ranked relatively higher in Table \ref{instruction-following-for-each-factor}, while all bottom three constraints have less than 90\% of the instruction-following score and are ranked relatively lower.
This suggests that our hypothesis is reasonable to some extent, not only across all constraints but also for each constraint.

Table \ref{tab:constrained_generation} showcases example essays generated by GPT-4 via instruction without and with a constraint on ``Usage'', which is ``Your essay must utilize a professional-level vocabulary''.
Professional words tend to have low frequencies in a corpus.
Thus, we identify low-frequency words\footnote{We leverage the Wikipedia dump extracted on April 23, 2023: \url{https://github.com/IlyaSemenov/wikipedia-word-frequency}. We define a word whose number of occurrences in the corpus is below the average number of occurrences of all words.
} in each text and observe that the constrained text contains more low-frequency words than the plain text.
It implies that the constrained text might follow the constraint. 


\section{Related Work}
\paragraph{LLM-Generated Text Detection Algorithms} 
In this section, we briefly outline current LLM-generated text detection algorithms.
The detection algorithms are mainly divided into three groups: watermarking, statistical outlier approach, and supervised classifiers.
The watermarking embeds token-level markers into output texts that are hard to recognize by humans and utilizes the ratio of the markers in a text for detection \cite{kirchenbauer2023watermark}. 
Our work only focuses on non-watermarked LLMs that are mainly for our daily use.
The statistical outlier approaches capture a probability deviation of a text from the predicted distribution of LLMs. These include token log probabilities \cite{solaiman2019release}, entropy \cite{ent08}, perplexity \cite{bere16}, and negative curvature of perturbed text probabilities \cite{mitchell2023detectgpt}.
The supervised classifiers are basically neural-based models trained to distinguish human-written and LLM-generated texts with labeled datasets \cite{uchendu-etal-2020-authorship,rodriguez-etal-2022-cross,guo2023close}. 
In addition to the above three groups, there has recently been a new direction: leveraging in-context learning for LLM-generated text detection \cite{Koike:OUTFOX:2024}. They utilize in-context learning of LLMs with retrieved few-shot human-written and LLM-generated examples, showing promising detection performance.

\paragraph{The Sensitivity of Prompting}
Prompting is a way of steering LLMs to generate texts via textual instruction without updating the model's parameters \citep{liu2021pretrain}.
Although prompting has shown promising performance on various tasks \citep{kamalloo2023evaluating,zhang2023benchmarking,zhang2023prompting}, the quality of output text is very sensitive to how the instruction is expressed \cite{jiang2020know}.
For instance, in machine translation, \citet{zhang2023prompting} observed that a small difference in generation instruction causes a significant difference in BLEU score of 23.1 points. 

Regardless of the substantial effect of instruction patterns on text quality, most studies on LLM detection overlook the subsequent effect of instruction patterns in text generation on the detection performance.
Our work investigates the impact of instruction patterns, encompassing not only the surface patterns but also the difference of task-oriented constraints in the instruction, on LLM detection.

\paragraph{Benchmarking datasets for LLM Detection}
Many studies recently have established benchmarking datasets to identify LLM-generated texts.
As representing examples, \citet{guo2023close} targets question answering and builds the Human ChatGPT Comparison Corpus (HC3) dataset for identifying ChatGPT-generated texts on diverse domains.
\citet{liu2023argugpt} focuses on argumentative essay writing and creates a corpus consisting of about 4,000 pairs of human-written and LLM-generated essays.
Highlighting how LLM-generated texts in such benchmarking datasets are generated, most studies make LLMs generate texts with one fixed instruction pattern. For instance, \citet{liu2023argugpt} targets on one instruction pattern: ``\texttt{\{essay\_topic\}} Do you agree or disagree? Use specific reasons and examples to support your answer. Write an essay of roughly \texttt{\{n\}} words.''.

Considering the above sensitivity of prompting, there is a gap between the instruction pattern for generation and LLM detection on the generated texts.
Our work bridges this gap by quantifying the effect of the difference in instructions on detection performance and showing a significant impact of the difference of task-oriented constraints in instructions.

\section{Conclusion}
This study investigates how much impact even task-oriented constraints in instruction can have on the current detector's performance to the generated texts.
Our experiments in the domain of student essay writing demonstrate that even task-oriented constraints in instruction have a more significant effect on the detection performance than the effect of sampling texts and paraphrasing the instruction.
Furthermore, there is an overall trend where the constraints can make LLM detection more challenging than without them.
Our analysis suggests that the high instruction-following ability of an LLM as a generator leads to a noteworthy effect of the constraint.

Taking into account the remarkable speed of recent LLM development, the instruction-following ability of LLMs would be much better, amplifying the effects of the constraint.
Therefore, in an era of evolving LLMs, our finding more strongly calls for further development of robust LLM detectors against such distribution shifts caused by a constraint in instruction.

\section*{Limitations}
Our work shows that even task-oriented constraints in generation instruction cause existing detectors to have a large variance in detection performance. We focus on student essay writing because 1) There is a recognized demand for LLM detection against academic dishonesty \cite{educatorconsiderations4chatgpt} with less discussion of such demand in other domains, 2) Due to the nature of being graded, the student essay domain has more established criteria we can refer to create the constraints than other domains (e.g., scientific writing and story generation).
Establishing such criteria for other domains could be another line of research, and constraints can vary from the criteria. Thus, we encourage the research community to further investigate the impact of constraints in other generation tasks on LLM detection.

\section*{Ethical Considerations}
Our goal in this paper is not to propose a method to deceive detectors.
Instead, we aim to improve the robustness of LLM-generated text detection and raise awareness in the research community that how the generation instruction is written has a large impact on detection performance. Furthermore, we provoke the research community to develop new robust LLM detectors against distribution shifts caused by constraints in generation instruction.

\section*{Acknowledgements}
These research results were obtained from the commissioned research (No.22501) by National Institute of Information and Communications Technology (NICT), Japan. In addition, this work was supported by JST SPRING, Japan Grant Number JPMJSP2106.
The authors are greatly thankful to Mengsay Loem, Marco Cognetta, and Youmi Ma for their detailed feedback on this paper.

\bibliography{anthology,custom}

\appendix

\section{Configuration Details}
\label{config_details}
\paragraph{Parameter Configurations of Generators}
For the essay generator models, we set the $\mathsf{temperature}$ parameter of ChatGPT to be 1.3, GPT-4 to be 1.0, and Davinci-002 to be 0.6.
For the paraphraser to rephrase the instruction in the \texttt{Paraprhase} setting, we set the $\mathsf{temperature}$ parameter of ChatGPT to be 1.3.

\paragraph{Details of the ICL Approach}
Following the setting of \citet{Koike:OUTFOX:2024}, we leverage ChatGPT (gpt-3.5-turbo-0613) as a detector of the ICL approach.
To eliminate the randomness of the detection, we configure $\mathsf{temperature}$ and $\mathsf{top\_p}$ parameters of ChatGPT to be 0.
As a dataset for retrieving examples for the ICL approach, we employ the training set\footnote{\url{https://github.com/ryuryukke/OUTFOX}} of \citet{Koike:OUTFOX:2024}, containing 14,400 triplets of essay problem statements, human-written essays\footnote{Written by native English speaking 6th to 12th graders from the U.S.}, and ChatGPT-generated essays.
Regardless of the type of essay generators (ChatGPT, GPT-4, and Davinci-002), we retrieve ChatGPT-generated essays.

\begin{table*}[t]
    \centering
    \small
    \setlength{\tabcolsep}{6pt} 
    \renewcommand{\arraystretch}{1.31} 
    \begin{tabular}{l}
    \hline
        Please compose a \{n\}-word essay based on the provided problem statement.\\
        I kindly request you to compose an essay that adheres to the given problem statement, ensuring that it contains \{n\} words.\\
        Could you kindly compose an essay containing \{n\} words based on the provided problem statement?\\
        Please compose an essay of \{n\} words based on the given prompt.\\
        Please compose an essay with a word count of \{n\}, based on the provided problem statement.\\
        Please compose an essay consisting of \{n\} words based on the provided problem statement.\\
        I kindly request you to compose an essay with \{n\} words, based on the subsequent problem statement.\\
        I kindly request you to compose an \{n\}-word essay based on the aforementioned problem statement.\\
        Please compose an essay of \{n\} words based on the provided problem statement.\\
        I am requesting an essay to be written in \{n\} words using the provided problem statement.\\
        Please compose an essay in which you discuss the given problem statement, utilizing \{n\} to express your thoughts.\\
        I kindly request you to compose an essay consisting of \{n\} words, using the problem statement provided below.\\
    \hline
    \end{tabular}
    \caption{Examples of the paraphrased instructions in the \texttt{paraphrase} setting.}
    \label{paraphrased_instructions}
\end{table*}

\paragraph{Examples of Paraphrased Instructions}
In the \texttt{paraphrase} setting, we employ ChatGPT to paraphrase the beginning of the original instruction, which is ``Given the following problem statement, please write an essay in \texttt{\{n\}} words.'' Table \ref{paraphrased_instructions} lists the paraphrased instructions.

\section{Do the Constraints Make Detection Easier or Harder?}
\label{appendix_b}
Our study mainly focuses on the SD of detection performance to elucidate the behavior of current significant detectors against task-oriented constraints in generation instruction.
This is because there is a common understanding that when building robust NLP systems, the performance of the system itself should be consistent, regardless of the scale of their performance \cite{yu-etal-2022-measuring}. Similarly, an ideal LLM detector should also have a consistent detection performance against the effect of task-oriented constraints, regardless of whether the performance improves or degrades.

On the other hand, we also acknowledge the worth of discussing whether LLMs can generate texts easier or harder to detect via instruction with the constraints.
Table \ref{detection-performance-for-each-constraint} showcases the detection performances of LLMs, including ChatGPT, GPT-4, and Davinci-002, with and without each constraint (Plain) in the generation instruction.
In most constraints, it is observed that the detection performance degrades (in gray parts) compared to the setting of Plain.
We can also see up to a 40.3 F1-score drop in blue parts with the lowest detection performance.
Finally, overall, there is a greater decrease in detection performance among ChatGPT and GPT-4 with relatively high instruction-following abilities compared to Davinci-002.
It suggests that instruction-following on the constraints could lead to not only higher detection performance deviation but also \textbf{more challenging }detection.

\begin{table*}[t]
    \centering
    \small
    \setlength{\tabcolsep}{3.8pt} 
    \renewcommand{\arraystretch}{1.35} 
    \begin{tabular}{cccccccccccccccc}
    \hline
     \multirow{2}{*}{\textbf{Generator}} & \multirow{2}{*}{\textbf{Detector}} & & \multicolumn{11}{c}{\textbf{Factor}} & \\\cline{4-14}
     & & \textbf{Plain} & \textbf{Grm.} & \textbf{Usg.} & \textbf{Mec.} & \textbf{Sty.} & \textbf{Rel.} & \textbf{Org.} & \textbf{Dev.} & \textbf{Chs.} & \textbf{Chr.} & \textbf{TC.} & \textbf{Per.} & \textbf{Avg.} & \textbf{Diff.} \\\hline
     \multirow{3}{*}{ChatGPT} & \multicolumn{1}{l}{HC3} & 78.2 & \cellcolor{gray!25} 78.1 & \cellcolor{blue!25} 37.9 & \cellcolor{gray!25} 74.9 & \cellcolor{gray!25} 56.8 & \cellcolor{gray!25} 77.8 & 78.9 & \cellcolor{gray!25} 76.0 & 83.9 & 79.7 & \cellcolor{gray!25} 76.7 & \cellcolor{gray!25} 73.2 & 72.2 & -6.03 \\
     & \multicolumn{1}{l}{ArguGPT} & 96.4 & \cellcolor{gray!25} 96.2 & \cellcolor{blue!25} 74.0 & \cellcolor{gray!25} 95.1 & \cellcolor{gray!25} 88.1 & \cellcolor{gray!25} 96.3 & 96.6 & 97.1 & 97.2 & 96.6 & \cellcolor{gray!25} 96.2 & \cellcolor{gray!25} 95.4 & 93.5 & -2.87 \\
     & \multicolumn{1}{l}{ICL} & 94.3 & \cellcolor{gray!25} 94.2 & 95.8 & \cellcolor{gray!25} 93.1 & 94.9 & 94.3 & 95.1 & \cellcolor{gray!25} 92.1 & 94.7 & \cellcolor{gray!25} 94.1 & \cellcolor{blue!25} 92.0 & 94.3 & 94.1 & -0.25 \\\hline
     \multirow{3}{*}{GPT-4} & \multicolumn{1}{l}{HC3} & 12.3 & \cellcolor{gray!25} 11.3 & \cellcolor{blue!25} 0.70 & \cellcolor{gray!25} 12.0 & \cellcolor{gray!25} 1.50 & \cellcolor{gray!25} 11.0 & 13.3 & \cellcolor{gray!25} 8.60 & 12.6 & \cellcolor{gray!25} 7.60 & \cellcolor{gray!25} 9.30 & \cellcolor{gray!25} 8.30 & 8.75 & -3.55 \\
     & \multicolumn{1}{l}{ArguGPT} & 84.0 & \cellcolor{gray!25} 83.4 & \cellcolor{blue!25} 40.9 & \cellcolor{gray!25} 83.3 & \cellcolor{gray!25} 50.9 & \cellcolor{gray!25} 82.3 & \cellcolor{gray!25} 82.3 & 84.9 & \cellcolor{gray!25} 80.1 & \cellcolor{gray!25} 70.4 & \cellcolor{gray!25} 81.5 & \cellcolor{gray!25} 79.5 & 74.5 & -9.50 \\
     & \multicolumn{1}{l}{ICL} & 92.2 & 92.6 & 93.5 & \cellcolor{gray!25} 91.3 & 92.3 & \cellcolor{gray!25} 91.6 & \cellcolor{gray!25} 91.7 & \cellcolor{gray!25} 92.1 & \cellcolor{gray!25} 91.7 & 92.8 & \cellcolor{gray!25} 90.7 & \cellcolor{blue!25} 88.5 & 91.7 & -0.49\\\hline
     \multirow{3}{*}{Davinci} & \multicolumn{1}{l}{HC3} & 87.2 & 90.1 & 87.7 & 87.9 & 89.2 & \cellcolor{blue!25} 85.4 & 89.8 & 88.4 & 87.9 & 89.6 & \cellcolor{gray!25} 86.4 & \cellcolor{gray!25} 87.0 & 88.1 & 0.93\\
     & \multicolumn{1}{l}{ArguGPT} & 97.8 & \cellcolor{gray!25} 97.6 & \cellcolor{gray!25} 97.3 & 97.8 & \cellcolor{gray!25} 97.6 & \cellcolor{gray!25} 97.2 & 97.5 & 97.4 & 97.7 & 97.7 & \cellcolor{blue!25} 96.7 & 97.2 & 97.4 & -0.37\\
     & \multicolumn{1}{l}{ICL} & 87.9 & \cellcolor{gray!25} 85.8 & 88.8 & \cellcolor{gray!25} 87.4 & \cellcolor{blue!25} 85.1 & \cellcolor{gray!25} 86.5 & 89.0 & \cellcolor{gray!25} 86.7 & 87.9 & 88.0 & \cellcolor{gray!25} 87.8 & 88.1 & 87.4 & -0.53\\\hline
    \end{tabular}
    \caption{A comparison of detection performance on essays generated via instructions with and without task-oriented constraints (Plain). \colorbox[gray]{0.8}{The gray parts} indicate a lower detection performance in the setting with the constraints than the setting of Plain. It depicts the lowest detection performance for each combination of generator and detector in \colorbox{blue!25}{the blue parts}.
    Avg. indicates the mean detection performance in the setting with the constraints. Diff. implies the difference in detection performance between Avg. and Plain.}
    \label{detection-performance-for-each-constraint}
\end{table*}

\begin{figure*}[t]
 \begin{center}
  \centering\includegraphics[width=\textwidth]{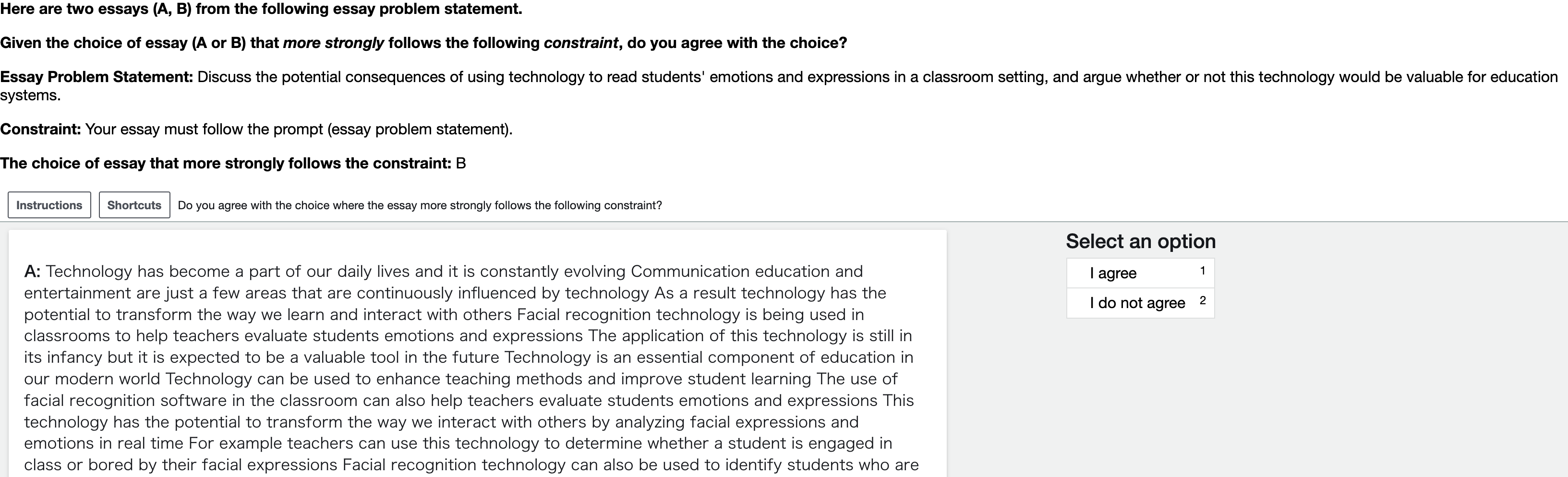}
  \caption{AMT interface for our human agreement test.}
  \label{fig:mturk}
 \end{center}
\end{figure*}

\section{Validation of GPT-4 Evaluation}
\label{human_agreement_details}
In \S\ref{analysis_gpt4}, we leverage GPT-4 to evaluate the instruction-following ability of LLMs.
Specifically, we instruct GPT-4 to decide whether the constrained text follows the constraint compared with the plain text. Here, the plain text and the constrained text are generated based on the same essay problem statement.

To ensure the validity of the GPT-4 evaluation, we utilize Amazon Mechanical Turk (AMT) to examine the ratio of the decisions made by GPT-4 that align with human consensus.
Figure \ref{fig:mturk} shows the AMT interface we use for our human agreement test.
Particularly, we show a constraint, an essay problem statement, the shuffled pair of plain and constrained texts based on the problem statement, and the GPT-4 decision to the AMT workers and ask if they agree with the decision.
We perform the test on 495 $\left(=45\times11\right)$ texts, which is a mixture of essays generated by ChatGPT and GPT-4 with the instruction including 11 constraints each.
We set workers' qualifications where the HIT\footnote{A human intelligence task (HIT) in AMT workplace refers to one single task. In our case, the HIT would be to decide if they agree or not with one of the GPT-4 decisions.} approval rate is over 99 \%, and the number of approved HITs is greater than 10,000.
In our test, one worker is assigned per HIT, and workers are paid \$0.03 per HIT.

\section{Computational Budget}
We run all the experiments on AI Bridging Cloud Infrastructure (ABCI)\footnote{\url{https://abci.ai/}}, Compute Node (V), whose CPUs are two Intel Xeon Gold 6148, and GPUs are four NVIDIA V100 SXM2. The total processing time is approximately 20 hours.

\end{document}